\documentclass[nonacm,sigconf]{acmart}
\AtBeginDocument{%
  }

\usepackage{times}
\usepackage{soul}
\usepackage{seqsplit}
\usepackage{url}

\usepackage{amssymb}

\usepackage[utf8]{inputenc}
\usepackage[font=small]{caption}
\usepackage{graphicx}
\usepackage{amsmath}
\usepackage{enumitem}
\usepackage{booktabs}
\usepackage[switch]{lineno}
\hypersetup{hidelinks}
\usepackage{amsmath,amssymb}
\usepackage{xcolor}

\DeclareUnicodeCharacter{0391}{\ensuremath{\mathrm{A}}}
\DeclareUnicodeCharacter{03B1}{\ensuremath{\alpha}}
\DeclareUnicodeCharacter{03B8}{\ensuremath{\theta}}
\DeclareUnicodeCharacter{03AE}{\ensuremath{\acute{\eta}}}
\DeclareUnicodeCharacter{03BD}{\ensuremath{\nu}}
\DeclareUnicodeCharacter{03B5}{\ensuremath{\varepsilon}}
\DeclareUnicodeCharacter{03B9}{\ensuremath{\iota}}
\DeclareUnicodeCharacter{03BA}{\ensuremath{\kappa}}
\DeclareUnicodeCharacter{03C1}{\ensuremath{\rho}}
\DeclareUnicodeCharacter{03C2}{\ensuremath{\varsigma}}
\DeclareUnicodeCharacter{03C5}{\ensuremath{\upsilon}}
\DeclareUnicodeCharacter{03B4}{\ensuremath{\delta}}
\DeclareUnicodeCharacter{03C0}{\ensuremath{\pi}}
\DeclareUnicodeCharacter{0395}{\ensuremath{\mathrm{E}}}
\DeclareUnicodeCharacter{0398}{\ensuremath{\Theta}}
\DeclareUnicodeCharacter{03AC}{\ensuremath{\acute{\alpha}}}
\DeclareUnicodeCharacter{03AF}{\ensuremath{\acute{\iota}}}
\DeclareUnicodeCharacter{03CC}{\ensuremath{\acute{o}}}
\DeclareUnicodeCharacter{03AD}{\ensuremath{\acute{\epsilon}}}
\DeclareUnicodeCharacter{03C4}{\ensuremath{\tau}}
\DeclareUnicodeCharacter{03BF}{\ensuremath{o}}
\DeclareUnicodeCharacter{03C3}{\ensuremath{\sigma}}
\DeclareUnicodeCharacter{03BC}{\ensuremath{\mu}}
\DeclareUnicodeCharacter{03BB}{\ensuremath{\lambda}}
\DeclareUnicodeCharacter{03C6}{\ensuremath{\varphi}}
\DeclareUnicodeCharacter{03C7}{\ensuremath{\chi}}

\setcopyright{acmlicensed}
\copyrightyear{2026}
\acmYear{2026}
\acmDOI{10.1145/3799682.3840068}

\acmConference[CIKM '26]{The 35th ACM International Conference on Information and Knowledge Management}{November 07--11, 2026}{Rome, Italy}

\acmISBN{978-1-4503-XXXX-X/2018/06}

\copyrightyear{2026}
\acmYear{2026}
\setcopyright{cc}
\setcctype{by-nc-nd}
\acmConference[CIKM '26]{Proceedings of the 35th ACM International Conference on Information and Knowledge Management}{November 07--11, 2026}{Rome, Italy}
\acmBooktitle{Proceedings of the 35th ACM International Conference on Information and Knowledge Management (CIKM '26), November 07--11, 2026, Rome, Italy}
\acmDOI{10.1145/3799682.3840068}
\acmISBN{979-8-4007-2539-5/2026/11}

\begin{document}

\title{	Automated Construction of FAIR Digital Object Knowledge Graphs from Flat Cultural Heritage Records}


\author{Zeyd Boukhers}
\correspondingauthor
\email{zeyd.boukhers@fit.fraunhofer.de}
\affiliation{%
  \institution{Fraunhofer Institute for Applied Information Technology FIT}
  \country{Germany}
}
\affiliation{%
  \institution{University Hospital of Cologne}
  \city{Cologne}
  \country{Germany}
}

\author{Lingxiao Kong}
\email{lingxiao.kong@fit.fraunhofer.de}
\affiliation{%
  \institution{Fraunhofer Institute for Applied Information Technology FIT}
  \country{Germany}
}
\affiliation{%
  \institution{University of Cologne}
  \city{Cologne}
  \country{Germany}
}

\author{Xenophon Zabulis}
\email{zabulis@ics.forth.gr}
\affiliation{%
  \institution{Foundation for Research and Technology}
  \country{Greece}
}

\author{Georgios Toubekis}
\email{georgios.toubekis@fit.fraunhofer.de}
\affiliation{%
  \institution{Fraunhofer Institute for Applied Information Technology FIT}
  \country{Germany}
}

\renewcommand{\shortauthors}{Boukhers et al.}

\begin{abstract}
The FAIR Digital Object (FDO) framework mandates that metadata attribute
values be expressed as persistent identifiers (PIDs) wherever possible, to produce a fully machine-actionable graph in which every reference is
resolvable. The Europeana Data Model was designed long before the FDO specification, and it stores most metadata values as plain text. This serves human browsing well enough, but gives an automated agent nothing to follow across records or collections. We present a pipeline that transforms flat Europeana records into an
FDO-compliant knowledge graph structured with CIDOC-CRM. Following the
FDO specification, we model every heritage entity as a discrete FDO with its own PID,
type, profile, and metadata layer. The core technical challenge is
automating the FDO-prescribed distinction between values that \emph{must}
become PID references (resolvable entities) and those that \emph{may}
remain literals (terminal leaves such as notes, measurements, and dates).
We address this with a large language model that classifies each metadata
value, routes it to a controlled vocabulary (Getty~AAT, Wikidata, VIAF, PeriodO),
and links it to a shared entity FDO. We evaluate using 637 archaeological records from five Europeana providers, processing each with the LLM. The pipeline links 86\% of metadata slots, resolving 58.5\% of values Europeana had not already enriched. It also merges cross-lingual surface forms that byte-identical matching keeps apart, where 17 of 33 such merges are correct on manual review. Graph connectivity does not separate this from string matching; what distinguishes the FDO graph is that every node is typed and resolvable.
\end{abstract}



\maketitle

\section{Introduction}

The FAIR Digital Object (FDO) framework~\cite{bonino2023fdo} provides an 
operational architecture for making digital resources Findable, 
Accessible, Interoperable, and Reusable by machines without human 
mediation. A core requirement of the FDO specification is that 
\emph{metadata attribute values are expressed as persistent identifiers 
(PIDs) rather than plain strings}~\cite{toubekis2026accountable}. The goal is to produce a fully machine-actionable graph in which every reference can be resolved, 
typed, and traversed algorithmically. Only terminal leaf values (i.e. numeric 
measurements, free-text notes, dates, checksums) may remain as 
literals~\cite{boukhers2024arxive,zoubia2025fdo}.

Cultural heritage aggregators such as Europeana~\cite{europeana2012} 
hold millions of records, yet the majority of metadata values remain 
expressed as plain-text strings rather than PID references. This is a gap 
between current practice and the degree of machine-actionability that the FDO 
framework envisions. A record's \texttt{dcterms:spatial} field typically
contains the string ``Bracara Augusta'' rather than a PID resolving
to a Place FDO.
A \texttt{dcCreator} field stores ``Cristina Braga'' rather than a
VIAF or ORCID reference. Machine actionability is therefore limited and no automated agent can
traverse from an object to its findspot or relate all objects from the same
excavation. It cannot, for example, tell that ``Braga'' and ``Bracara Augusta'' name
the same place.

To maximise machine actionability in compliance with the FDO 
specification, metadata values that denote reusable entities must be 
replaced by PID references to shared FDOs. The technical challenge, 
then, is not merely one of entity linking. It is rather the automated 
enforcement of the FDO-prescribed boundary between \emph{PID-expressed 
values} (entities that can and should be shared across records) and 
\emph{literal-expressed values} (terminal leaves that cannot be 
meaningfully shared). This boundary is defined by the FDO architecture 
itself~\cite{boukhers2024arxive}: a value is a PID reference if and 
only if it denotes a reusable real-world entity (place, material, 
actor, event, concept). Otherwise, it remains a literal.

We present a pipeline that operationalises this principle at scale. 
Given flat Europeana records, a large language model (LLM) classifies 
each metadata value into a semantic type and determines whether it
crosses the PID/literal boundary. Values that cross it are resolved to a
shared entity FDO identified in a controlled vocabulary. The 
resulting knowledge graph follows the FDO architecture 
of~\cite{boukhers2024arxive,zoubia2025fdo}: every entity is a discrete FDO with its 
own persistent identifier, FDO Type, FDO Profile, and Metadata FDO 
carrying CIDOC-CRM~\cite{cidoccrm} assertions. The contribution here lies in automating the population of these FDOs from flat records rather than redefining the architecture itself.

Our contributions are:
\begin{enumerate}[leftmargin=*]
  \item An LLM-driven pipeline that automates the FDO-compliant
  transformation of flat metadata into PID references, achieving
  58.5\% resolution on values it must resolve itself and 89.6\%
  overall once Europeana's pre-linked URIs are counted, reported separately
  throughout.
  \item An FDO-native knowledge graph instantiated automatically by our pipeline, in which every 
  resolved entity is a full FDO (i.e. self-describing, PID-resolvable, and 
  carrying declared operations), building directly on the FDO Manager 
  specification~\cite{zoubia2025fdo}.
  \item An empirical evaluation on 637 records from five providers, with
every record processed by the LLM. Over an identical slot population,
entity resolution reduces disconnected components from 32 to 20; given
all content values, including those the pipeline leaves unlinked, string
matching reaches 8. The direction of the connectivity result depends on
the slot population, which bounds what connectivity establishes about
machine actionability (Section~\ref{sec:eval}).
\end{enumerate}

\section{Related Work}\label{sec:related}

\paragraph{FAIR Digital Objects and machine actionability.}
The FAIR principles shifted data-management practice from human-readable documentation toward resources that can be discovered, accessed, interpreted, and reused by machines~\cite{wilkinson2016fair}. FAIR Digital Objects (FDOs) operationalise this goal by treating digital resources as persistent, typed, and self-describing units with machine-resolvable identifiers and explicit metadata~\cite{desmedt2020fdo,bonino2023fdo}. Recent FDO work further emphasises the separation between a minimal PID/kernel layer and richer semantic descriptions, so that machines can first resolve an object, inspect its type and profile, and then retrieve domain-specific metadata or operations~\cite{zoubia2025fdo}. Autonomous FDOs add policy and agreement layers so that objects validate and reconcile their own assertions~\cite{boukhers2026autonomous}, which builds on the resolvable metadata values we produce here. We adopt this architecture, but address a different problem: given flat cultural-heritage records, how can the system decide which metadata values should become PID references to reusable entities and which values should remain literals? This PID/literal boundary is central to producing machine-actionable FDO graphs, yet it is not addressed by existing FDO infrastructure alone.

\paragraph{Cultural heritage metadata, CIDOC-CRM, and linked data.}
Cultural heritage institutions have long used semantic models to improve interoperability across heterogeneous collections. The Europeana Data Model (EDM) provides an aggregation-oriented model for publishing and enriching collection records~\cite{europeana2012}, while CIDOC-CRM provides an event-centric ontology for integrating complex heritage information across institutions~\cite{cidoccrm}. Prior work has shown that CIDOC-CRM and EDM can capture complementary aspects of museum data, but also that real institutional records remain difficult to model because they are heterogeneous, incomplete, historically accumulated, and often expressed as weakly structured strings~\cite{dijkshoorn2018modeling}. Cultural heritage linked-data projects such as the Zeri Photo Archive, the Smithsonian American Art Museum, and the Beyond 2022 knowledge graph demonstrate the value of RDF, CIDOC-CRM, named graphs, and external authority links for collection discovery and scholarly interpretation~\cite{daquino2017zeri,szekely2013saam,debruyne2022ireland}. However, these efforts typically rely on substantial manual modelling, project-specific mappings, or curated transformation rules. In contrast, our pipeline targets aggregator-scale Europeana records and automates the conversion from flat metadata slots into FDO entities and CIDOC-CRM relations.

\paragraph{Entity linking and cultural-heritage authority control.}
Entity linking connects textual mentions to canonical entities in knowledge bases and has been widely studied as a core step for semantic search, knowledge-base population, and data integration~\cite{shen2015entity,sevgili2022neural}. Wikidata has become especially important because it is multilingual, continuously updated, community-curated, and broadly connected to external authority files~\cite{vrandecic2023wikidata,moller2022wikidata}. In the cultural heritage domain, systems such as Heritage Connector use machine learning to build linked open data from museum catalogues, perform record linkage to Wikidata, and extract new entities from textual collection descriptions~\cite{dutia2021heritage}. More recent cultural-heritage knowledge-graph work combines entity extraction, relation extraction, and deep learning to structure fragmented museum data~\cite{huang2023heritagekg}. These approaches are close to our goal, but they optimise primarily for linking or graph construction. Our task is narrower and more FDO-specific: before linking, the system must determine whether a metadata value denotes a reusable entity that should be represented by a PID, or a terminal attribute that should remain a literal. This makes our problem not only entity linking, but specification-driven metadata normalisation for machine-actionable FDO construction.

\paragraph{AI-based semantic enrichment for cultural heritage.}
A recent line of work enriches cultural resources with AI-generated
identifications, entity links, and graph structure. CulturAI applies
semantic technologies and machine learning to augment cultural records with
external knowledge~\cite{carta2022culturai}; comparative evaluations of
state-of-the-art entity linkers on community-generated cultural content
report that off-the-shelf models degrade on domain vocabulary and
multilingual mentions~\cite{benkhedda2024enriching}; multimodal systems
extend heritage knowledge graphs by combining language and vision
models~\cite{zhang2026multimodal}; and LLMs combined with ontological
engineering have been used to generate knowledge graphs from
cultural heritage texts~\cite{schimmenti2025kgg}. These systems target
richer or more accurate links than we do, and several are stronger
entity linkers. What differs here is the decision the system is asked to
make. Those pipelines link mentions that are given to them as entities; ours
must first decide \emph{which} values are entities at all, because the FDO
specification, not a benchmark, fixes what may remain a literal. Our
contribution is that boundary decision and its use to populate an FDO
graph, rather than linking accuracy, which we do not benchmark against
dedicated linkers.

\paragraph{LLMs for data integration and knowledge-graph construction.}
Recent data-management research shows that foundation models can generalise to classical data tasks such as schema matching, data cleaning, and entity resolution when these tasks are formulated through prompts or structured outputs~\cite{narayan2022foundation,li2024llmdata}. In parallel, work on the convergence of large language models and knowledge graphs argues that LLMs can support KG construction, completion, and semantic enrichment, while KGs provide grounding, structure, and interpretability~\cite{pan2024llmkg}. General KG surveys also identify entity canonicalisation, type assignment, relation extraction, and knowledge fusion as persistent challenges in automatic graph construction~\cite{hogan2021kg,zhong2023akgc}. Our pipeline applies this emerging LLM--KG paradigm to cultural heritage metadata, but with an explicit architectural constraint from the FDO model: the LLM is not used merely to generate triples, but to enforce the PID/literal decision, route resolvable values to appropriate authority vocabularies, and instantiate the resulting entities as FDOs connected through CIDOC-CRM properties. This positions the work between cultural heritage linked data, entity linking, and FDO-based machine actionability.
\section{System Overview}

\begin{figure*}[t]
  \centering
  \includegraphics[width=0.75\linewidth]{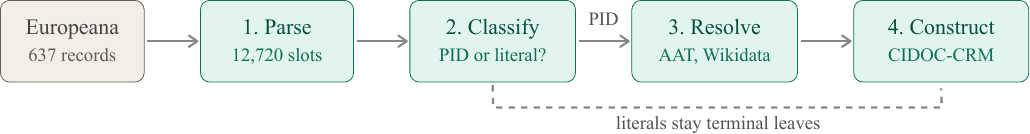}
  \caption{Pipeline architecture. Flat Europeana records are parsed into 
  (field, value) slots. An LLM enforces the FDO-prescribed PID/literal 
  boundary by classifying each value's semantic type and resolvability. 
  Resolvable values are linked to shared entity FDOs via controlled 
  vocabularies; literals remain as terminal leaves.}
  \label{fig:pipeline}
\end{figure*}

Our pipeline transforms flat Europeana records into an FDO-compliant 
knowledge graph through four stages (Figure~\ref{fig:pipeline}), following 
the architecture of~\cite{boukhers2024arxive,zoubia2025fdo}. The 
design is governed such that: 1) \emph{every heritage entity is a discrete 
FDO; 2) metadata values are PIDs wherever the value denotes a reusable 
entity; 3) literals are reserved only for terminal leaves}.

\paragraph{Stage 1: Slot Extraction and Prior Assignment:}

Each Europeana record is retrieved from the Search API under the
\texttt{rich} profile and decomposed into a set of \emph{slots} $(f, v)$
where $f$ is the metadata field and $v$ is its value. This projection
carries EDM-derived fields rather than raw EDM elements, so the field names
are those of the API response: \texttt{dctermsSpatial}, \texttt{edmPlace},
\texttt{dcCreator}, \texttt{edmConcept}, \texttt{edmTimespan} and
\texttt{organizations} for entity-bearing values, and \texttt{title},
\texttt{dcDescription}, \texttt{rights}, \texttt{edmIsShownAt},\\
\texttt{dcLanguage} and \texttt{year} for values expected to remain
literal. Each field carries a \emph{prior semantic class}
(e.g., \texttt{dctermsSpatial}~$\allowbreak\, \to$~PLACE,
\texttt{edmTimespan}~$\to$~PERIOD) that the LLM may override. Neither
\texttt{dcterms:medium} nor \texttt{dc:type} occurs in this projection,
which is why the corpus yields almost no MATERIAL slots
(Section~\ref{sec:eval}). Slots whose values are already 
URIs from Europeana's enrichment pipeline are classified directly from 
the URI domain (Getty TGN~$\to$~PLACE, AAT~$\to$~OBJECT\_TYPE, 
ULAN~$\to$~ACTOR) and bypass the LLM.

\paragraph{Stage 2: PID/Literal Boundary Classification}

For each non-pre-linked slot, we must determine whether $v$ denotes a 
reusable entity that should become a PID reference, or a terminal leaf 
that remains literal. Per the FDO specification~\cite{boukhers2024arxive}, 
the distinction is:
\begin{itemize}[leftmargin=*]
  \item \textbf{PID reference}: the value denotes a place, material, 
  actor, period, event, object type, or other shared concept that can 
  be identified in a controlled vocabulary and reused across records.
  \item \textbf{Literal}: the value is a free-text description, numeric 
  measurement, date, access URL, inventory number, or rights statement 
  that cannot be meaningfully shared as a discrete entity.
\end{itemize}

We frame this as a structured classification task for an LLM. Given 
$(f, v)$ and the record's context (title, type, country), the model 
outputs: (1)~a semantic type from a 16-class taxonomy; 
(2)~a binary \emph{resolvable} decision implementing the PID/literal 
boundary; and (3)~the target vocabulary for resolution 
(AAT, Wikidata, VIAF, PeriodO). All three decisions are
produced in a single structured JSON call.

\paragraph{Stage 3: Entity Resolution and Linking}

For each slot classified as resolvable, the pipeline queries the 
target vocabulary's API using an LLM-generated search string. A 
vocabulary fallback chain ensures robustness:
VIAF~$\to$~Wikidata, AAT~$\to$~Wikidata,
PeriodO~$\to$~AAT~$\to$~Wikidata. PLACE values route to Wikidata:
GeoNames is not used, as no endpoint credential was available.
When multiple
candidates are returned with similar confidence scores, the LLM
reranks them using the record context for disambiguation.

\paragraph{Stage 4: FDO Graph Construction}

Following the architecture of~\cite{boukhers2024arxive} and the 
kernel/metadata separation of~\cite{zoubia2025fdo}, resolved entities 
are instantiated as full FDOs:

\textbf{Kernel Record.} A lean structural envelope containing only 
PIDs: this FDO's PID, FDO Type PID, FDO Profile PID, Metadata FDO PID, 
Content PIDs, applicable Operation PIDs, a timestamp, and a checksum. 
No domain semantics reside in the kernel.

\textbf{Metadata FDO.} A separate FDO carrying CIDOC-CRM property 
assertions. Object FDOs (E22\_Human-Made\_Object) link to entity FDOs 
via CRM properties: \texttt{P53\_has\_former\_or\\\_current\_location} 
points to a Place FDO (E53),\\ \texttt{P45\_consists\_of} to a Material
FDO (E57),\\ \texttt{P14\_carried\_out\_by} to a Person FDO (E21) and
\texttt{\seqsplit{P50\_has\_current\_keeper}} to a Group FDO (E74), separating the
actor who acted from the institution that holds the object;
\texttt{P2\_has\_type} to a Type FDO (E55); and
\texttt{P10\_falls\_within} to a Period FDO (E4).
Crucially, property objects are PIDs of other FDOs---not vocabulary URIs 
directly. External URIs (e.g., \texttt{\seqsplit{http://vocab.getty.edu/tgn/\ldots}}) 
appear only as \texttt{owl:sameAs} on the entity FDO itself.

\textbf{Self-describing infrastructure.} FDO Types (Place, Material, 
Actor, etc.) and FDO Operations (Resolve, GetMetadata, Validate, Link) 
are themselves FDOs with PIDs, registered in a type/operations registry. 
This makes the graph navigable by machines without out-of-band 
documentation~\cite{zoubia2025fdo}.

The result is a knowledge graph where every node is a PID-resolvable 
FDO, every metadata edge is a typed CIDOC-CRM relation, and every 
entity can be traversed, queried, and composed by automated agents. 
Whether this constitutes machine actionability in the sense the FDO 
specification intends is what Section~\ref{sec:eval} sets out to test.

\section{Evaluation}
\label{sec:eval}
We evaluate two research questions:
(RQ1)~What does graph connectivity capture, and what does it miss?
(RQ2)~How consistently does the pipeline type and route the values it
resolves?

\paragraph{Dataset and Setup:} We harvest 637 archaeological records from five Europeana data
providers spanning four countries, yielding
12,720 metadata slots over content-bearing fields (see below);
9,197 already carry a vocabulary URI from Europeana's enrichment and
3,523 are plain strings.
The LLM is a 120B-parameter instruction-tuned model (``gpt-oss-120b'') run at
temperature~0 with every response cached, which makes re-runs deterministic. Every one of the 637 records is processed by the LLM. Code, configuration, and
evaluation outputs are available at
\url{https://github.com/ZResearch/aFDO_CIKM}.
The accompanying test suite covers the record parser, the FDO kernel and
serialisation layer, and the graph metrics.

\textbf{Field restriction.} All figures below are computed over
content-bearing fields. Five fields are excluded, for two different
reasons. \texttt{organizations} (six distinct aggregator identifiers
across all 637 records), \texttt{rights} and \texttt{edmIsShownAt} carry
provenance rather than content. \texttt{dcLanguage} is excluded because
the classifier fails on it: language codes are marked resolvable and
linked to unrelated art-historical concepts---``en'' and ``English'' both
resolve to the AAT concept \emph{English Baroque} across 334 records, and
``ro'' to \emph{Romanian (Cyrillic)} across 148. These pass the
type--vocabulary check, which validates a URI's domain rather than the
entity behind it. They are 28\% of pipeline links and are a boundary
classification failure, not a routing failure; we report them here and
exclude them from the figures below.

\paragraph{RQ1: What Connectivity Captures.}

We operationalise machine actionability as graph
connectivity---a machine agent can traverse the graph from one record to
another if and only if they share at least one node. We compare
$G_\text{flat}$, records connected by byte-identical string values appearing
in $\geq$2 records, against $G_\text{FDO}$, records connected through shared
entity FDOs.

Two controls are required. First, both graphs must be built from the
same slot population, since a baseline that excludes URI-valued slots that the
FDO graph counts is weaker by construction. Second, connectivity saturates.
A single Creative Commons rights URI appears in 589 of 637 records and
connects the flat graph on its own; the aggregator \texttt{organization}
URIs do the same to the FDO graph. With those hubs present, all three graphs collapse to a single component and the metric distinguishes
nothing. We therefore restrict both graphs to content-bearing fields and
drop any value occurring in more than 10\% of records.

\begin{table}[t]
\centering
\caption{Connectivity over 637 records, content-bearing
fields, values occurring in more than 10\% of records removed.
$G_\text{flat}^{=}$ is restricted to the same slots as $G_\text{FDO}$;
$G_\text{flat}^{+}$ adds the content values the pipeline left unlinked, which
still connect records as byte-identical strings.}
\label{tab:connectivity}
\small

\begin{tabular}{lrrr}
\toprule
\textbf{Metric} & $G_\text{flat}^{+}$ & $G_\text{flat}^{=}$ & $G_\text{FDO}$ \\
\midrule
Slots used & 12720 & 10995 & 10995 \\
Distinct shared nodes & 479 & 359 & 369 \\
Connected components & 8 & 32 & \textbf{20} \\
Largest component (\% records) & 71.6\% & 69.4\% & \textbf{73.6\%} \\
Cross-provider bridges & 13 & 12 & 13 \\
\midrule
Nodes with resolvable URI & 63.8\% & 63.8\% & 100.0\%$^\dagger$ \\
Surface forms merged & 0 & 0 & \textbf{176} \\
\midrule
\multicolumn{4}{l}{\textit{Slot-level resolution}} \\
\quad Pre-linked URIs & \multicolumn{3}{r}{9197/9197 = 100.0\%$^\dagger$} \\
\quad Resolved by pipeline & \multicolumn{3}{r}{1798/3075 = 58.5\%} \\
\quad Combined & \multicolumn{3}{r}{10995/12272 = 89.6\%} \\
\bottomrule
\multicolumn{4}{l}{\footnotesize $^\dagger$ induced by construction, not measured.}
\end{tabular}
\end{table}

Table~\ref{tab:connectivity} reports the result. Given the same 10,995
slots, entity resolution reduces components from 32 to 20 and raises the
largest component from 69.4\% to 73.6\%. Given all 12,720 content values,
string matching reaches 8 components: values the pipeline leaves unlinked
still connect records as byte-identical strings. Connectivity therefore does
not separate the two approaches in a stable way, and its direction depends on
a choice of slot population that the metric itself does not fix. Europeana
already ships vocabulary URIs for 65\% of slots, and those URIs are
byte-identical across records, so string matching recovers them without any
resolution step. Over content-bearing fields the pipeline produces 800 entity
FDOs and 9,054 CIDOC-CRM edges.

\textbf{Surface-form merging.} Connectivity counts reachable records; it
does not distinguish a node that is a resolved entity from one that is a
shared string. Two properties do. Every node in $G_\text{FDO}$ carries a
resolvable external identifier and a semantic type, against 63.8\% of
$G_\text{flat}$ nodes that happen to be URI strings, the remainder being bare
literals that support no further resolution. And 33 entity FDOs absorb 176
distinct surface forms that byte-identical matching keeps apart.

We reviewed all 33 merges by hand: 17 are correct, and 16 are not. The
correct ones absorb 34 surface forms but attach to 327 records, and are
cross-lingual or variant alignments of places and organisations---Athens/%
``Αθήνα'' across 135 records, National Theatre of Greece/``Εθνικό Θέατρο''
across 29, Epidaurus/``Επίδαυρος'' across 12. The incorrect ones absorb 142
forms across only 163 records and concentrate in PERIOD: 59 internal record
identifiers of the form \texttt{share3d:*/TMP.1} resolve to a concept
labelled \emph{Human parainfluenza virus 1} and 47 more to
\emph{ISO 2-numbers}. Failures are numerous but shallow; successes are few
but deep. PERIOD is also the only class where type--vocabulary consistency is
measurable, at 62.5\%, and all 124 observed type errors fall in it.

\paragraph{RQ2: Typing and Routing Consistency.}

\begin{table}[t]
\centering
\caption{Entity resolution quality, content-bearing
fields. Pre-linked slots are omitted from the accuracy rows: their semantic
type is inferred from the same URI domain the check validates, so that the figure
measures the URI rules rather than the model.}
\label{tab:quality}
\small
\begin{tabular}{lr}
\toprule
\textbf{Metric} & \textbf{Value} \\
\midrule
Total linked slots & 10995 \\
\quad Pre-linked URIs (Europeana) & 9197 \\
\quad Resolved by pipeline & 1798 \\
\midrule
\multicolumn{2}{l}{\textit{Type--vocabulary consistency (pipeline-resolved)}} \\
\quad PERIOD ($n$=331) & 62.5\% \\
\quad Other classes & not measurable \\
\midrule
\multicolumn{2}{l}{\textit{Vocabulary routing (pipeline-resolved)}} \\
\quad Strict ($n$=1798) & 73.1\% \\
\midrule
\multicolumn{2}{l}{\textit{Surface-form merging}} \\
\quad Entities merging $\geq$2 forms & 33 \\
\quad Correct on manual review & 17 \\
\bottomrule
\end{tabular}
\end{table}

\textbf{Label similarity is not reported.} Fuzzy string overlap between a
source value and its resolved label is a common proxy for link quality, but
it is invalid on a multilingual corpus: it scores \emph{correct} cross-lingual
links at zero. ``Αθήνα''~$\to$~``Athens'' and ``Επίδαυρος''~$\to$~``Epidaurus''
are right and score 0.00. Over this corpus, the mean is 0.493 with 43.6\% of
links below 0.3, and the low scores are dominated by correct Greek and
Romanian resolutions. The metric penalises precisely the alignments that
distinguish resolution from string matching, so we report no figure for it.

\textbf{Typing and routing.} Type--vocabulary consistency is reported only
for pipeline-resolved slots, since for pre-linked slots the type is inferred
from the URI domain the check then validates. It is measurable for PERIOD
alone, at 62.5\% on 331 slots; the other classes route predominantly to
Wikidata, where any semantic type is admissible, and the check is vacuous.
Strict vocabulary routing is 73.1\% (1315/1798). A fallback-inclusive figure
can also be computed, but it counts any fall-through to Wikidata as correct
and is therefore bounded by the design rather than measured, so we omit it.

\paragraph{Discussion.}\label{sec:discussion}

\textbf{What connectivity measures.} Connectivity is a weak discriminator
here. It saturates on values that denote no entity---a rights statement
shared by 589 records connects the graph as effectively as a resolved
place---and its verdict reverses with the choice of slot population. What
separates the graphs is not how many records are reachable but what the nodes
are: typed, resolvable identifiers rather than strings. The denominators are
as follows: of 12,720 slots the classifier marks 12,272 resolvable and 10,995
of those are linked, giving 89.6\%. That denominator is the classifier's own
decision, so a value wrongly ruled literal is invisible to this measure.

\textbf{Cross-provider bridges.}
Thirteen entity FDOs are shared across providers.
A machine agent
starting at a record in the National Heritage Institute of Bucharest can
reach records in the Huis van Hilde collection (Netherlands) through a shared
\emph{Type} FDO (AAT 300192802); no Place FDO connects those two providers,
and all Place bridges in the corpus are intra-Romanian. One of the 13 is spurious---an AAT concept labelled
``Human parainfluenza virus 1'' typed as PERIOD, spanning 59 records across
two countries. Twelve of the 13 carry a raw
URI as their label, because pre-linked URIs are not dereferenced for
canonical labels; this limits how far such bridges can be audited
automatically.

\textbf{Limitations.} The current evaluation uses automatic quality
proxies rather than human-annotated gold; a full annotation study is
planned.
This is the main limitation, and it bounds every number above: none of
these measures verifies that a link is the right entity, that a literal was
correctly left unresolved, or that a bridge is meaningful.
Type--vocabulary consistency is agreement with a URI-domain rule, not
with a human judgement, and it is measurable for one class only.
Two further limits are specific to this corpus. The Search API projection
carries no material or medium field, so the pipeline produces
no MATERIAL entities and the \texttt{P45\_consists\_of} relation is
never instantiated; the edge distribution is dominated by
\texttt{P2\_has\_type} (4701) and \texttt{P53} (4658), with
\texttt{P10} (1278), \texttt{P14} (245) and \texttt{P50} (113).
And all results come from one domain, archaeology, in
one aggregator, so we make no claim of generalisation to other collections
or aggregators.

\section{Conclusion}

We presented an LLM-driven pipeline for transforming flat Europeana metadata into a machine-actionable FAIR Digital Object knowledge graph. The central contribution is the automated enforcement of the FDO PID/literal boundary. Specifically, values that denote reusable cultural heritage entities are resolved to persistent identifiers, while terminal attributes remain literals. By representing resolved
places, actors, periods, object types, and concepts
as full FDOs connected through CIDOC-CRM relations, the resulting graph moves beyond conventional metadata enrichment toward a self-describing structure that machines can resolve, traverse, and reuse.

This work shows that FDO principles can be operationalised over heterogeneous aggregator metadata without requiring fully manual modelling.
The measured benefit is that nodes are typed, resolvable, and merge surface variants, not that more records are reachable. It also rests on automatic proxies rather than human judgement.
More broadly, it suggests that LLMs can support the transition from string-based cultural heritage records to interoperable, PID-centred knowledge infrastructures. Future work will
prioritise a human-annotated gold set, which is the prerequisite for
every quality claim above; compare against dedicated entity-linking and
record-linkage systems; replace graph connectivity with a downstream
traversal or discovery task, since connectivity saturates on trivial shared
values; and extend the pipeline to additional Europeana domains and
aggregators to test whether any of this generalises.

\newpage
\section*{Acknowledgements}
This work was supported by the European Union's Horizon Europe programme
through the ARXIVE project (Grant Agreement No.~101233418) and the ARGUS
project (Grant Agreement \\ No.~101132308), and by the German Research
Foundation (DFG) through NFDI4DataScience (NFDI4DS)
(project no.~460234259).

\section*{GenAI Usage Disclosure}
The authors used generative AI tools to assist with language editing, paragraph restructuring, and drafting of non-technical explanatory text. The authors also used generative AI tools to support software development, including assistance with code drafting, debugging, and improving implementation clarity. All generated text and code were reviewed, verified, tested where applicable, and approved by the authors. The authors take full responsibility for the content, technical claims, implementation, results, and conclusions presented in this submission.

\bibliographystyle{ACM-Reference-Format}
\bibliography{sample-base}

\end{document}